\documentclass[11pt]{article}
\usepackage{graphicx,amsmath,amsbsy,amssymb,citesort,epsfig,epsf}

\def\real    { \mathbb{R} }

\newcommand{\qed}{{\unskip\nobreak\hfil\penalty50\hskip2em\vadjust{}
           \nobreak\hfil$\Box$\parfillskip=0pt\finalhyphendemerits=0\par}}
\newtheorem{thm}{Theorem}

\newtheorem{definition}{Definition}

\newcommand{\eps}{\epsilon}
\newcommand{\bitem}{\begin{itemize}}
\newcommand{\eitem}{\end{itemize}}

\newcommand{\beqn}{\begin{equation}}
\newcommand{\eeqn}{\end{equation}}
\newcommand{\balign}{\begin{align}}
\newcommand{\ealign}{\end{align}}

\def \RipConst {C_0}
\def \CandesOne {C_1}
\def \CandesTwo {C_2}
\def \CddOne {C_3}
\def \CddTwo {C_4}
\def \CddThree {C_5}
\def \KappaOne {C_6}
\def \KappaTwo {C_7}
\def \GrimesOne {C_8}
\def \GrimesTwo {C_9}
\def \GrimesThree {C_{10}}

\def \proj { {\Phi} } 
\def \noise { \eta } 
\def \dim {N}      
\def \pdim {M}     

\def \manifold { \mathcal{M} } 
\def \condition { \tau } 
\def \regularity { R } 
\def \volume { V } 
\def \gdist { {d_\manifold} } 
\def \sres { T } 

\def \sparsity {K} 
\def \mdim {K}     

\newcommand{\dist}[2]{\left\| #1 - #2 \right\|_2}
\newcommand{\norm}[1]{\left\| #1 \right\|_2}

\newcommand{\compactionB}[3]{(1 - #3)#2 \le #1 \le (1 +
#3)#2}

\title{{\bf Manifold-Based Signal Recovery and Parameter Estimation from Compressive Measurements}}

\author{Michael B. Wakin\footnote{Division of Engineering, Colorado
School of Mines. Email: mwakin@mines.edu. This research was
partially supported by NSF Grant DMS-0603606 and DARPA Grant
HR0011-08-1-0078. The content of this article does not necessarily
reflect the position or the policy of the Government and no official
endorsement should be inferred.}}

\begin{document}

\maketitle

\vspace{-0.2in}

\begin{abstract}
A field known as Compressive Sensing (CS) has recently emerged to
help address the growing challenges of capturing and processing
high-dimensional signals and data sets. CS exploits the surprising
fact that the information contained in a sparse signal can be
preserved in a small number of compressive (or random) linear
measurements of that signal. Strong theoretical guarantees have been
established on the accuracy to which sparse or near-sparse signals
can be recovered from noisy compressive measurements. In this paper,
we address similar questions in the context of a different modeling
framework. Instead of sparse models, we focus on the broad class of
manifold models, which can arise in both parametric and
non-parametric signal families. Building upon recent results
concerning the stable embeddings of manifolds within the measurement
space, we establish both deterministic and probabilistic
instance-optimal bounds in $\ell_2$ for manifold-based signal
recovery and parameter estimation from noisy compressive
measurements. In line with analogous results for sparsity-based CS,
we conclude that much stronger bounds are possible in the
probabilistic setting. Our work supports the growing empirical
evidence that manifold-based models can be used with high accuracy
in compressive signal processing.
\end{abstract}

\noindent {\bf Keywords.} Manifolds, dimensionality reduction,
random projections, Compressive Sensing, sparsity, signal recovery,
parameter estimation, Johnson-Lindenstrauss lemma.

~

\noindent {\bf AMS Subject Classification.} 53A07, 57R40, 62H12,
68P30, 94A12, 94A29.

\section{Introduction}
\label{sec:intro}

\subsection{Concise signal models}

A significant byproduct of the Information Age has been an explosion
in the sheer quantity of raw data demanded from sensing systems.
From digital cameras to mobile devices, scientific computing to
medical imaging, and remote surveillance to signals intelligence,
the size (or dimension) $\dim$ of a typical desired signal continues
to increase. Naturally, the dimension $\dim$ imposes a direct burden
on the various stages of the data processing pipeline, from the data
acquisition itself to the subsequent transmission, storage, and/or
analysis; and despite rapid and continual improvements in computer
processing power, other bottlenecks do remain, such as communication
bandwidth over wireless channels, battery power in remote sensors
and handheld devices, and the resolution/bandwidth of
analog-to-digital converters.

Fortunately, in many cases, the information contained within a
high-dimensional signal actually obeys some sort of concise,
low-dimensional model. Such a signal may be described as having just
$\sparsity \ll \dim$ degrees of freedom for some $\sparsity$.
Periodic signals bandlimited to a certain frequency are one example;
they live along a fixed $\sparsity$-dimensional linear subspace of
$\real^\dim$. Piecewise smooth signals are an example of {\em sparse
signals}, which can be written as a succinct linear combination of
just $\sparsity$ elements from some basis such as a wavelet
dictionary. Still other signals may live along $\mdim$-dimensional
submanifolds of the ambient signal space $\real^\dim$; examples
include collections of signals observed from multiple viewpoints in
a camera or sensor network. In general, the conciseness of these
models suggests the possibility for efficient processing and
compression of these signals.

\subsection{Compressive measurements}

Recently, the conciseness of certain signal models has led to the
use of {\em compressive measurements} for simplifying the data
acquisition process. Rather than designing a sensor to measure a
signal $x \in \real^\dim$, for example, it often suffices to design
a sensor that can measure a much shorter vector $y = \proj x$, where
$\proj$ is a linear measurement operator represented as an $\pdim
\times \dim$ matrix, and where typically $\pdim \ll \dim$. As we
discuss below in the context of Compressive Sensing (CS), when
$\proj$ is properly designed, the requisite number of measurements
$\pdim$ typically scales with the information level $\sparsity$ of
the signal, rather than with its ambient dimension $\dim$.

Surprisingly, the requirements on the measurement matrix $\proj$ can
often be met by choosing $\proj$ randomly from an acceptable
distribution. One distribution allows the entries of $\proj$ to be
chosen as i.i.d.\ Gaussian random variables; another dictates that
$\proj$ has orthogonal rows that span a random $\pdim$-dimensional
subspace of $\real^\dim$.

Physical architectures have been proposed for hardware that will
enable the acquisition of signals using compressive
measurements~\cite{duarte2008spi,candes2008ics,healy2008cpi,demod}.
The potential benefits for data acquisition are numerous. These
systems can enable simple, low-cost acquisition of a signal directly
in compressed form without requiring knowledge of the signal
structure in advance. Some of the many possible applications include
distributed source coding in sensor networks~\cite{dcsJournal},
medical imaging~\cite{lustig2008csm}, high-rate analog-to-digital
conversion~\cite{candes2008ics,healy2008cpi,demod}, and error
control coding~\cite{CandesDLP}.

\subsection{Signal understanding from compressive measurements}

Having acquired a signal $x$ in compressed form (in the form of a
measurement vector $y$), there are many questions that may then be
asked of the signal. These include:
\begin{itemize}
\item [Q1.] {\em Recovery:} What was the original signal $x$?
\item [Q2.] {\em Sketching:} Supposing that $x$ was sparse or nearly so,
what were the $\sparsity$ basis vectors used to generate $x$?
\item [Q3.] {\em Parameter estimation:} Supposing $x$ was generated from a
$\mdim$-dimensional parametric model, what was the original
$\mdim$-dimensional parameter that generated $x$?
\end{itemize}
Given only the measurements $y$ (possibly corrupted by noise),
solving any of the above problems requires exploiting the concise,
$\sparsity$-dimensional structure inherent in the
signal.\footnote{Other problems, such as finding the nearest
neighbor to $x$ in a large database of signals~\cite{JL_Indyk}, can
also be solved using compressive measurements and do not require
assumptions about the concise structure in $x$.} CS addresses
questions Q1 and Q2 under the assumption that the signal $x$ is
$\sparsity$-sparse (or approximately so) in some basis or
dictionary; in Section~\ref{sec:cs} we outline several key
theoretical bounds from CS regarding the accuracy to which these
questions may be answered.

\subsection{Manifold models for signal understanding}
\label{sec:mmodelsunder}

In this paper, we will address these questions in the context of a
different modeling framework for concise signal structure. Instead
of sparse models, we focus on the broad class of {\em manifold
models}, which arise both in settings where a $\mdim$-dimensional
parameter $\theta$ controls the generation of the signal and also in
non-parametric settings.

\begin{figure}[t]
\begin{center}
\epsfysize = 48mm \epsffile{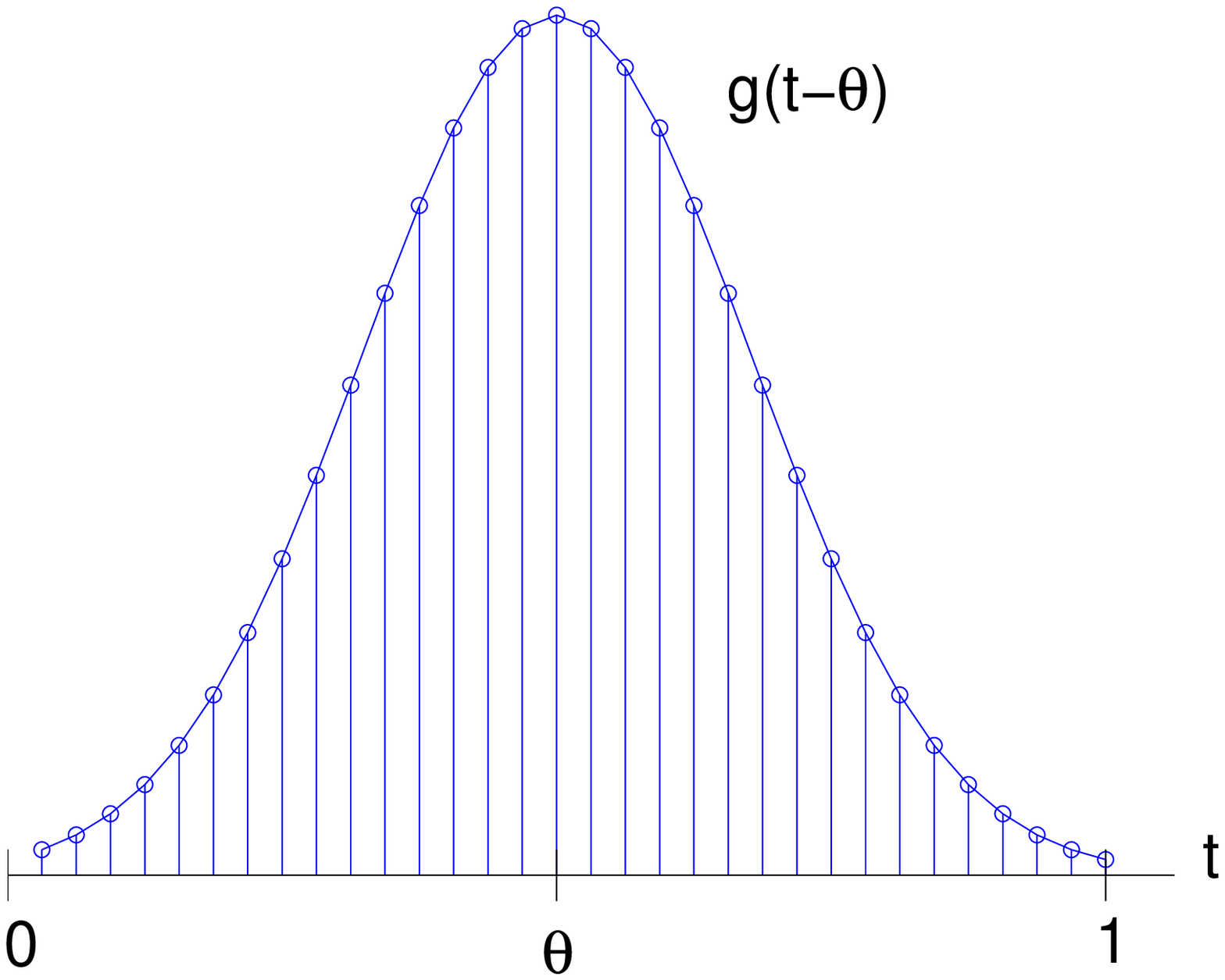} \quad\quad
\epsfysize = 48mm \epsffile{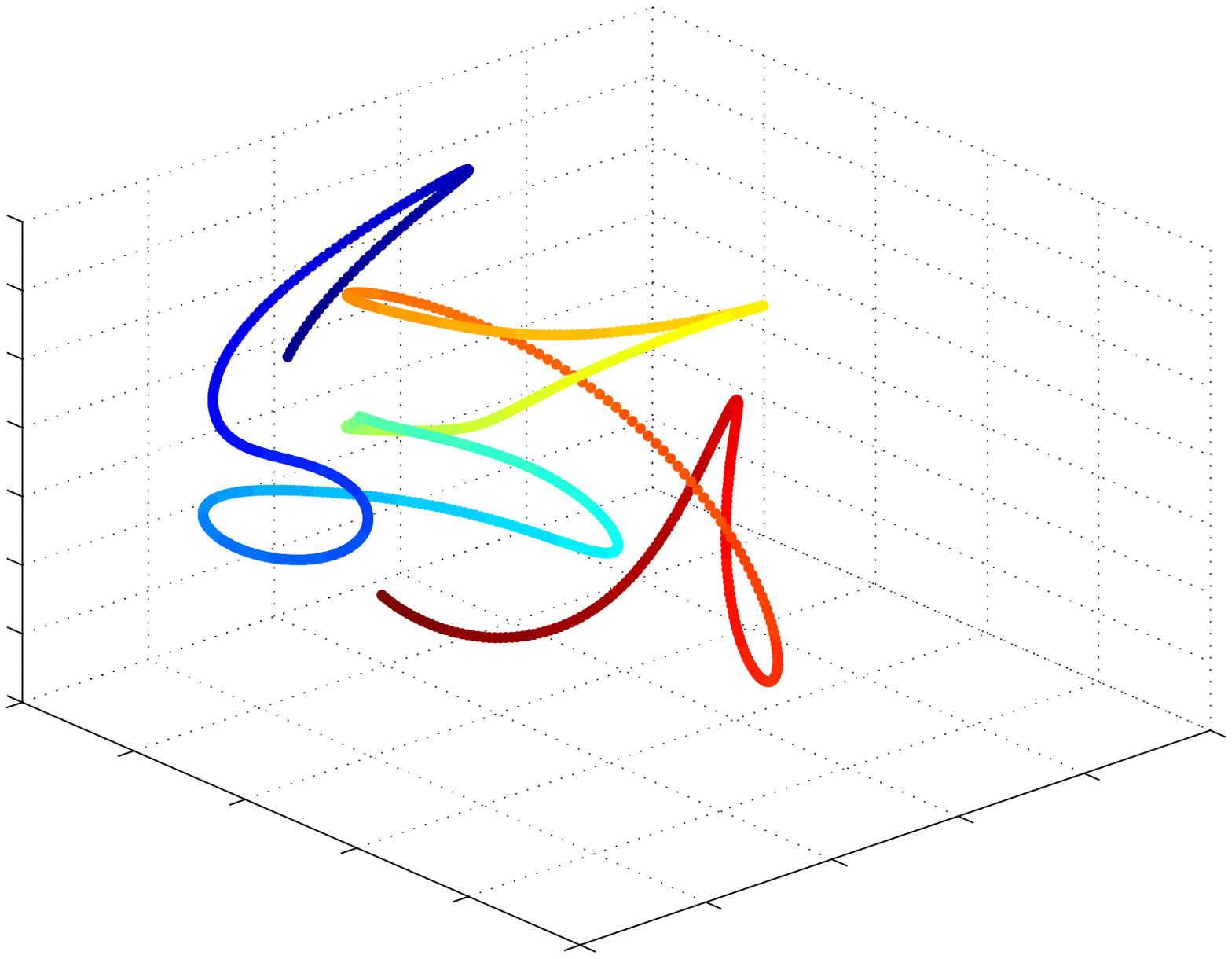} \quad\quad\quad
\end{center}
\vspace*{-5mm} \caption{\small\sl \label{fig:CSM} (a) The
articulated signal $f_\theta(t) = g(t-\theta)$ is defined via shifts
of a primitive function $g$, where $g$ is a Gaussian pulse. Each
signal is sampled at $\dim$ points, and as $\theta$ changes, the
resulting signals trace out a 1-D manifold in $\real^\dim$. (b)
Projection of the manifold from $\real^\dim$ onto a random 3-D
subspace; the color/shading represents different values of $\theta
\in [0,1]$.}
\end{figure}

As a very simple illustration, consider the articulated signal in
Figure~\ref{fig:CSM}(a). We let $g(t)$ be a fixed continuous-time
Gaussian pulse centered at $t=0$ and consider a shifted version of
$g$ denoted as the parametric signal $f_\theta(t) := g(t-\theta)$
with $t,\theta \in [0,1]$. We then suppose the discrete-time signal
$x = x_\theta \in \real^\dim$ arises by sampling the continuous-time
signal $f_\theta(t)$ uniformly in time, i.e., $x_\theta(n) =
f_\theta(n/\dim)$ for $n=1,2,\dots,\dim$. As the parameter $\theta$
changes, the signals $x_\theta$ trace out a continuous
one-dimensional (1-D) curve $\manifold = \{x_\theta: \theta \in
[0,1]\} \subset \real^\dim$. The conciseness of our model (in
contrast with the potentially high dimension $\dim$ of the signal
space) is reflected in the low dimension of the path $\manifold$.

In the real world, manifold models may arise in a variety of
settings. A $\mdim$-dimensional parameter $\theta$ could reflect
uncertainty about the 1-D timing of the arrival of a signal (as in
Figure~\ref{fig:CSM}(a)), the 2-D orientation and position of an
edge in an image, the 2-D translation of an image under study, the
multiple degrees of freedom in positioning a camera or sensor to
measure a scene, the physical degrees of freedom in an articulated
robotic or sensing system, or combinations of the above. Manifolds
have also been proposed as approximate models for signal databases
such as collections of images of human faces or of handwritten
digits~\cite{Eigenfaces,digits,broomhead01wh}.

Consequently, the potential applications of manifold models are
numerous in signal processing. In some applications, the signal $x$
itself may be the object of interest, and the concise manifold model
may facilitate the acquisition or compression of that signal.
Alternatively, in parametric settings one may be interested in using
a signal $x = x_\theta$ to infer the parameter $\theta$ that
generated that signal.
In an application known as manifold learning, one may be presented
with a collection of data $\{x_{\theta_1}, x_{\theta_2}, \dots,
x_{\theta_n}\}$ sampled from a parametric manifold and wish to
discover the underlying parameterization that generated that
manifold.
Multiple manifolds can also be considered simultaneously, for
example in problems that require recognizing an object from one of
$n$ possible classes, where the viewpoint of the object is uncertain
during the image capture process. In this case, we may wish to know
which of $n$ manifolds is closest to the observed image $x$.

While any of these questions may be answered with full knowledge of
the high-dimensional signal $x \in \real^\dim$, there is growing
theoretical and experimental support that they can also be answered
from only compressive measurements $y = \proj x$. In a recent paper,
we have shown that given a sufficient number $\pdim$ of random
measurements, one can ensure with high probability that a manifold
$\manifold \subset \real^\dim$ has a stable embedding in the
measurement space $\real^\pdim$ under the operator $\proj$, such
that pairwise Euclidean and geodesic distances are approximately
preserved on its image $\proj \manifold$. We restate the precise
result in Section~\ref{sec:manifolds}, but a key aspect is that the
number of requisite measurements $\pdim$ is linearly proportional to
the information level of the signal, i.e., the dimension $\mdim$ of
the manifold.

As a very simple illustration of this embedding phenomenon,
Figure~\ref{fig:CSM}(b) presents an experiment where just $\pdim =
3$ compressive measurements are acquired from each point $x_\theta$
described in Figure~\ref{fig:CSM}(a). We let $\dim = 1024$ and
construct a randomly generated $3 \times \dim$ matrix $\proj$ with
orthogonal rows. Each point $x_\theta$ from the original manifold
$\manifold \subset \real^{1024}$ maps to a unique point $\proj
x_\theta$ in $\real^3$; the manifold embeds in the low-dimensional
measurement space. Given any $y = \proj x_{\theta'}$ for $\theta'$
unknown, then, it is possible to infer the value $\theta'$ using
only knowledge of the parametric model for $\manifold$ and the
measurement operator $\proj$. Moreover, as the number $\pdim$ of
compressive measurements increases, the manifold embedding becomes
much more stable and remains highly self-avoiding.

Indeed, there is strong empirical evidence that, as a consequence of
this phenomenon, questions such as Q1 (signal recovery) and Q3
(parameter estimation) can be accurately solved using only
compressive measurements of a signal $x$, and that these procedures
are robust to noise and to deviations of the signal $x$ away from
the manifold $\manifold$~\cite{mbwPhdThesis,davenport2007sfc}.
Additional theoretical and empirical justification has followed for
the manifold learning~\cite{hegde2007rpm} and multiclass recognition
problems~\cite{davenport2007sfc} described above.
Consequently, many of the advantages of compressive measurements
that are beneficial in sparsity-based CS (low-cost sensor design,
reduced transmission requirements, reduced storage requirements,
lack of need for advance knowledge of signal structure, simplified
computation in the low-dimensional space $\real^\pdim$, etc.)\ may
also be enjoyed in settings where manifold models capture the
concise signal structure. Moreover, the use of a manifold model can
often capture the structure of a signal in many fewer degrees of
freedom $\mdim$ than would be required in any sparse representation,
and thus the measurement rate $\pdim$ can be greatly reduced
compared to sparsity-based CS approaches.

In this paper, we will focus on questions Q1 (signal recovery) and
Q3 (parameter estimation) and reinforce the existing empirical work
by establishing theoretical bounds on the accuracy to which these
questions may be answered.
We will consider both deterministic and probabilistic
instance-optimal bounds, and we will see strong similarities to
analogous results that have been derived for sparsity-based CS. As
with sparsity-based CS, we show for manifold-based CS that for any
fixed $\proj$, uniform deterministic $\ell_2$ recovery bounds for
recovery of all $x$ are necessarily poor. We then show that, as with
sparsity-based CS, providing for any $x$ a probabilistic bound that
holds over most $\proj$ is possible with the desired accuracy. We
consider both noise-free and noisy measurement settings and compare
our bounds with sparsity-based CS.

\subsection{Paper organization}

We begin in Section~\ref{sec:cs} with a brief review of CS topics,
to set notation and to outline several key results for later
comparison. In Section~\ref{sec:manifolds} we discuss manifold
models in more depth, restate our previous bound regarding stable
embeddings of manifolds, and formalize our criteria for answering
questions Q1 and Q3 in the context of manifold models. In
Section~\ref{sec:deter}, we confront the task of deriving
deterministic instance-optimal bounds in $\ell_2$. In
Section~\ref{sec:prob}, we consider instead probabilistic
instance-optimal bounds in $\ell_2$. We conclude in
Section~\ref{sec:concl} with a final discussion.

\section{Sparsity-Based Compressive Sensing} \label{sec:cs}

\subsection{Sparse models}

The concise modeling framework used in Compressive Sensing (CS) is
{\em sparsity}. Consider a signal $x \in \real^\dim$ and suppose the
$\dim \times \dim$ matrix $\Psi = [\psi_1 ~ \psi_2 ~ \cdots~
\psi_N]$ forms an orthonormal basis for $\real^\dim$. We say $x$ is
$\sparsity$-sparse in the basis $\Psi$ if for $\alpha \in
\real^\dim$ we can write
$$
x = \Psi \alpha,
$$
where $\|\alpha\|_0 = \sparsity < \dim$. (The $\ell_0$-norm notation
counts the number of nonzeros of the entries of $\alpha$.) In a
sparse representation, the actual information content of a signal is
contained exclusively in the $\sparsity < \dim$ positions and values
of its nonzero coefficients.

For those signals that are approximately sparse, we may measure
their proximity to sparse signals as follows. We define
$\alpha_\sparsity \in \real^\dim$ to be the vector containing only
the largest $\sparsity$ entries of $\alpha$, with the remaining
entries set to zero. Similarly, we let $x_\sparsity = \Psi
\alpha_\sparsity$. It is then common to measure the proximity to
sparseness using either $\|\alpha-\alpha_\sparsity\|_1$ or
$\norm{\alpha-\alpha_\sparsity}$ (the latter of which equals
$\norm{x-x_\sparsity}$ because $\Psi$ is orthonormal).

\subsection{Compressive measurements}
\label{sec:csmeas}

CS uses the concept of sparsity to simplify the data acquisition
process. Rather than designing a sensor to measure a signal $x \in
\real^\dim$, for example, it often suffices to design a sensor that
can measure a much shorter vector $y = \proj x$, where $\proj$ is a
linear measurement operator represented as an $\pdim \times \dim$
matrix, and typically $\pdim \ll \dim$.

The measurement matrix $\proj$ must have certain properties in order
to be suitable for CS. One desirable property (which leads to the
theoretical results we mention in Section~\ref{sec:csresults}) is
known as the Restricted Isometry Property
(RIP)~\cite{CandesUES,CandesECLP,CandesSSR}. We say a matrix $\proj$
meets the {\em RIP of order $\sparsity$ with respect to the basis
$\Psi$} if for some $\delta_\sparsity > 0$,
$$
(1-\delta_\sparsity) \norm{\alpha} \le \norm{\proj \Psi \alpha} \le
(1+\delta_\sparsity) \norm{\alpha}
$$
holds for all $\alpha \in \real^\dim$ with $\|\alpha\|_0 \le
\sparsity$. Intuitively, the RIP can be viewed as guaranteeing a
{\em stable embedding} of the collection of $\sparsity$-sparse
signals within the measurement space $\real^\pdim$. In particular,
supposing the RIP of order $2\sparsity$ is satisfied with respect to
the basis $\Psi$, then for all pairs of $\sparsity$-sparse signals
$x_1, x_2 \in \real^\dim$, we have
\begin{equation}
(1-\delta_{2\sparsity}) \norm{x_1-x_2} \le \norm{\proj x_1 - \proj
x_2} \le (1+\delta_{2\sparsity}) \norm{x_1-x_2}. \label{eq:rip2}
\end{equation}

Although deterministic constructions of matrices meeting the RIP are
still a work in progress, it is known that the RIP often be met by
choosing $\proj$ randomly from an acceptable distribution. For
example, let $\Psi$ be a fixed orthonormal basis for $\real^\dim$
and suppose that
\begin{equation}
 \pdim \ge \RipConst \sparsity
\log(\dim/\sparsity) \label{eq:nummeas}
\end{equation}
for some constant $\RipConst$. Then supposing that the entries of
the $\pdim \times \dim$ matrix $\proj$ are drawn as independent,
identically distributed Gaussian random variables with mean $0$ and
variance $\frac{1}{\pdim}$, it follows that with high probability
$\proj$ meets the RIP of order $\sparsity$ with respect to the basis
$\Psi$. Two aspects of this construction deserve special notice:
first, the number $\pdim$ of measurements required is linearly
proportional to the information level $\sparsity$, and second,
neither the sparse basis $\Psi$ nor the locations of the nonzero
entries of $\alpha$ need be known when designing the measurement
operator $\proj$. Other random distributions for $\proj$ may also be
used, all requiring approximately the same number of measurements.
One of these distributions~\cite{dasgupta99elementary,JLCS} dictates
that $\proj = \sqrt{\dim/\pdim}\, \Xi$, where $\Xi$ is an $\pdim
\times \dim$ matrix having orthonormal rows that span a random
$\pdim$-dimensional subspace of $\real^\dim$. We refer to such
choice of $\proj$ as a {\em random orthoprojector}.\footnote{Our
previous use of the term ``random orthoprojector'' in~\cite{mbwFocm}
excluded the normalization factor of $\sqrt{\dim/\pdim}$. However we
find it more appropriate to include this factor in the current
paper.}

\subsection{Signal recovery and sketching}
\label{sec:csresults}

Although the sparse structure of a signal $x$ need not be known when
collecting measurements $y = \proj x$, a hallmark of CS is the use
of the sparse model in order to facilitate understanding from the
compressive measurements. A variety of algorithms have been proposed
to answer Q1 (signal recovery), where we seek to solve the
apparently undercomplete set of $\pdim$ linear equations $y = \proj
x$ for $\dim$ unknowns. The canonical
method~\cite{DonohoCS,CandesUES,CandesRUP} is known as {\em
$\ell_1$-minimization} and is formulated as follows: first solve
\begin{equation}
\widehat{\alpha} = \arg\min_{\alpha' \in \real^\dim} \|\alpha'\|_1
~\mathrm{subject~to}~ y = \proj \Psi \alpha', \label{eq:l1min}
\end{equation}
and then set $\widehat{x} = \Psi \widehat{\alpha}$. Under this
recovery program, the following bounds are known.

\begin{thm}{\em \cite{candes2008rip}}
Suppose that $\proj$ satisfies the RIP of order $2\sparsity$ with
respect to $\Psi$ and with constant $\delta_{2\sparsity} <
\sqrt{2}-1$. Let $x \in \real^\dim$, suppose $y = \proj x$, and let
the recovered estimates $\widehat{\alpha}$ and $\widehat{x}$ be as
defined above. Then
\begin{equation}
\norm{x-\widehat{x}} = \norm{\alpha-\widehat{\alpha}} \le \CandesOne
\sparsity^{-1/2} \|\alpha-\alpha_\sparsity\|_1 \label{eq:mixed1}
\end{equation}
for a constant $\CandesOne$. In particular, if $x$ is
$\sparsity$-sparse, then $\widehat{x} = x$. \label{theo:mixed1}
\end{thm}

This result can be extended to account for measurement noise.

\begin{thm}{\em \cite{candes2008rip}} Suppose that $\proj$
satisfies the RIP of order $2\sparsity$ with respect to $\Psi$ and
with constant $\delta_{2\sparsity} < \sqrt{2}-1$. Let $x \in
\real^\dim$, and suppose that
$$
y = \proj x + \noise
$$
where $\norm{\noise} \le \epsilon$. Then let
$$ \widehat{\alpha} =
\arg\min_{\alpha' \in \real^\dim} \|\alpha'\|_1
~\mathrm{subject~to}~ \norm{y - \proj \Psi \alpha'} \le \epsilon,
$$
and set $\widehat{x} = \Psi \widehat{\alpha}$. Then
\begin{equation}
\norm{x-\widehat{x}} = \norm{\alpha-\widehat{\alpha}} \le \CandesOne
\sparsity^{-1/2} \|\alpha-\alpha_\sparsity\|_1 + \CandesTwo
\epsilon. \label{eq:mixed2}
\end{equation}
for constants $\CandesOne$ (which is the same as above) and
$\CandesTwo$.
 \label{theo:mixed2}
\end{thm}

These results are not unique to $\ell_1$ minimization; similar
bounds have been established for signal recovery using greedy
iterative algorithms ROMP~\cite{romp} and CoSAMP~\cite{cosamp}.
Bounds of this type are extremely encouraging for signal processing.
From only $\pdim$ measurements, it is possible to recover $x$ with
quality that is comparable to its proximity to the nearest
$\sparsity$-sparse signal, and if $x$ itself is $\sparsity$-sparse
and there is no measurement noise, then $x$ can be recovered
exactly. Moreover, despite the apparent ill-conditioning of the
inverse problem, the measurement noise is not dramatically amplified
in the recovery process.

These bounds are known as {\em deterministic}, {\em
instance-optimal} bounds because they hold deterministically for any
$\proj$ that meets the RIP, and because for a given $\proj$ they
give a guarantee for recovery of any $x \in \real^\dim$ based on its
proximity to the concise model.

The use of $\ell_1$ as a measure for proximity to the concise model
(on the right hand side of (\ref{eq:mixed1}) and (\ref{eq:mixed2}))
arises due to the difficulty in establishing $\ell_2$ bounds on the
right hand side. Indeed, it is known that deterministic $\ell_2$
instance-optimal bounds cannot exist that are comparable to
(\ref{eq:mixed1}) and (\ref{eq:mixed2}). In particular, for any
$\proj$, to ensure that $\norm{x-\widehat{x}} \le \CddOne
\norm{x-x_\sparsity}$ for all $x$, it is known~\cite{CDD} that this
requires that $\pdim \ge \CddTwo \dim$ regardless of $\sparsity$.

However, it is possible to obtain an instance-optimal $\ell_2$ bound
for sparse signal recovery in the noise-free setting by changing
from a deterministic formulation to a {\em probabilistic}
one~\cite{CDD,devore08in}. In particular, by considering any given
$x \in \real^\dim$, it is possible to show that for {\em most}
random $\proj$, letting the measurements $y = \proj x$, and
recovering $\widehat{x}$ via $\ell_1$-minimization (\ref{eq:l1min}),
it holds that
\begin{equation}
\norm{x-\widehat{x}} \le \CddThree
\norm{x-x_\sparsity}.\label{eq:devore}
\end{equation}
While the proof of this statement~\cite{devore08in} does not involve
the RIP directly, it holds for many of the same random distributions
that work for RIP matrices, and it requires the same number of
measurements (\ref{eq:nummeas}) up to a constant.

Similar bounds hold for the closely related problem of Q2
(sketching), where the goal is to use the compressive measurement
vector $y$ to identify and report only approximately $\sparsity$
expansion coefficients that best describe the original signal, i.e.,
a sparse approximation to $\alpha_\sparsity$. In the case where
$\Psi = I$, an efficient randomized measurement process coupled with
a customized recovery algorithm~\cite{gilbert2007osa} provides
signal sketches that meet a deterministic mixed-norm $\ell_2/\ell_1$
instance-optimal bound analogous to~(\ref{eq:mixed1}). A desirable
aspect of this construction is that the computational complexity
scales with only $\log(\dim)$ (and is polynomial in $\sparsity$);
this is possible because only approximately $\sparsity$ pieces of
information must be computed to describe the signal. For signals
that are sparse in the Fourier domain ($\Psi$ consists of the DFT
vectors), probabilistic $\ell_2/\ell_2$ instance-optimal bounds have
also been established~\cite{gilbert05im} that are analogous to
(\ref{eq:devore}).

\section{Compressive Measurements of Manifold-Modeled Signals}
\label{sec:manifolds}

\subsection{Manifold models}
\label{sec:mmodels}

As we have discussed in Section~\ref{sec:mmodelsunder}, there are
many possible modeling frameworks for capturing concise signal
structure. Among these possibilities are the broad class of manifold
models.

Manifold models arise, for example, in settings where the signals of
interest vary continuously as a function of some $\mdim$-dimensional
parameter. Suppose, for instance, that there exists some parameter
$\theta$ that controls the generation of the signal. We let
$x_\theta \in \real^\dim$ denote the signal corresponding to the
parameter $\theta$, and we let $\Theta$ denote the
$\mdim$-dimensional parameter space from which $\theta$ is drawn. In
general, $\Theta$ itself may be a $\mdim$-dimensional manifold and
need not be embedded in an ambient Euclidean space. For example,
supposing $\theta$ describes the 1-D rotation parameter in a
top-down satellite image, we have $\Theta = S^1$.

Under certain conditions on the parameterization $\theta \mapsto
x_\theta$, it follows that
$$
\manifold := \{x_\theta: \theta \in \Theta\}
$$
forms a $\mdim$-dimensional {\em submanifold} of $\real^\dim$. An
appropriate visualization is that the set $\manifold$ forms a
nonlinear $\mdim$-dimensional ``surface'' within the
high-dimensional ambient signal space $\real^\dim$. Depending on the
circumstances, we may measure the distance between points two points
$x_{\theta_1}$ and $x_{\theta_2}$ on the manifold $\manifold$ using
either the ambient Euclidean distance
$$
\norm{x_{\theta_1}-x_{\theta_2}}
$$
or the geodesic distance along the manifold, which we denote as
$\gdist(x_{\theta_1},x_{\theta_2})$. In the case where the geodesic
distance along $\manifold$ equals the native distance in parameter
space, i.e., when
\begin{equation}
\gdist(x_{\theta_1},x_{\theta_2}) = d_\Theta(\theta_1, \theta_2),
\label{eq:isometry}
\end{equation}
we say that $\manifold$ is {\em isometric} to $\Theta$. The
definition of the distance $d_\Theta(\theta_1, \theta_2)$ depends on
the appropriate metric for the parameter space $\Theta$; supposing
$\Theta$ is a convex subset of Euclidean space, then we have
$d_\Theta(\theta_1, \theta_2) = \norm{\theta_1-\theta_2}$.

While our discussion above concentrates on the case of manifolds
$\manifold$ generated by underlying parameterizations, we stress
that manifolds have also been proposed as approximate
low-dimensional models within $\real^\dim$ for nonparametric signal
classes such as images of human faces or handwritten
digits~\cite{Eigenfaces,digits,broomhead01wh}. These signal families
may also be considered.

The results we present in this paper will make reference to certain
characteristic properties of the manifold under study. These terms
are originally defined in~\cite{niyogi,mbwFocm} and are repeated
here for completeness. First, our results will depend on a measure
of regularity for the manifold. For this purpose, we adopt the {\em
condition number} defined recently by Niyogi et al.~\cite{niyogi}.

\begin{definition}{\em \cite{niyogi}}
Let $\manifold$ be a compact Riemannian submanifold of $\real^\dim$.
The {\em condition number} is defined as $1/\condition$, where
$\condition$ is the largest number having the following property:
The open normal bundle about $\manifold$ of radius $r$ is embedded
in $\real^\dim$ for all $r < \condition$. \label{def:cn}
\end{definition}

The condition number $1/\condition$ controls both local properties
and global properties of the manifold. Its role is summarized in two
key relationships~\cite{niyogi}. First, the the curvature of any
unit-speed geodesic path on $\manifold$ is bounded by
$1/\condition$. Second, at long geodesic distances, the condition
number controls how close the manifold may curve back upon itself.
For example, supposing $x_1,x_2 \in \manifold$ with $\gdist(x_1,x_2)
> \condition$, it must hold that $\norm{x_1-x_2} > \condition/2$.

We also require a notion of ``geodesic covering regularity'' for a
manifold. While this property is not the focus of the present paper,
we include its definition in Appendix~\ref{app:reg} for
completeness.

We conclude with a brief but concrete example to illustrate specific
values for these quantities. Let $\dim
> 0$, $\kappa > 0$, $\Theta = \real \!\! \mod 2\pi$, and suppose $x_\theta
\in \real^\dim$ is given by
$$
x_\theta = [\kappa \cos(\theta); ~ \kappa \sin(\theta); ~ 0; ~0;
\cdots 0]^T.
$$
In this case, $\manifold = \{x_\theta: \theta \in \Theta\}$ forms a
circle of radius $\kappa$ in the $x(1), x(2)$ plane. The manifold
dimension $\mdim=1$, the condition number $\condition = \kappa$, and
the geodesic covering regularity $\regularity$ can be chosen as any
number larger than $\frac{1}{2}$. We also refer in our results to
the $\mdim$-dimensional volume $\volume$ of the $\manifold$, which
in this example corresponds to the circumference $2\pi \kappa$ of
the circle.

\subsection{Stable embeddings of manifolds}
\label{sec:mstable}

In cases where the signal class of interest $\manifold$ forms a
low-dimensional submanifold of $\real^\dim$, we have theoretical
justification that the information necessary to distinguish and
recover signals $x \in \manifold$ can be well-preserved under a
sufficient number of compressive measurements $y = \proj x$. In
particular, we have recently shown that an RIP-like property holds
for families of manifold-modeled signals.
\begin{thm} {\em \cite{mbwFocm}}
Let $\manifold$ be a compact $\mdim$-dimensional Riemannian
submanifold of $\real^\dim$ having condition number $1/\condition$,
volume $\volume$, and geodesic covering regularity $R$. Fix $0 <
\eps < 1$ and $0 < \rho < 1$. Let $\proj$ be a random $\pdim \times
\dim$ orthoprojector with
\begin{equation} \pdim = O\left(\frac{\mdim \log(\dim \volume \regularity  \condition^{-1} \eps^{-1}) \log(1/\rho)}{\eps^2}
\right).
\label{eq:mmeasmain}
\end{equation}
If $\pdim \le \dim$, then with probability at least $1-\rho$ the
following statement holds: For every pair of points $x_1,x_2 \in
\manifold$,
\begin{equation}
(1-\epsilon) \norm{x_1-x_2} \le \norm{\proj x_1 - \proj x_2} \le
(1+\epsilon) \norm{x_1-x_2}. \label{eq:mmain}
\end{equation}
\label{theo:manifoldjl}
\end{thm}

The proof of this theorem involves the Johnson-Lindenstrauss
Lemma~\cite{dasgupta99elementary,dfrp,JLCS}, which guarantees a
stable embedding for a finite point cloud under a sufficient number
of random projections. In essence, manifolds with higher volume or
with greater curvature have more complexity and require a more dense
covering for application of the Johnson-Lindenstrauss Lemma; this
leads to an increased number of measurements~(\ref{eq:mmeasmain}).

By comparing (\ref{eq:rip2}) with (\ref{eq:mmain}), we see a strong
analogy to the RIP of order $2\sparsity$. This theorem establishes
that, like the class of $\sparsity$-sparse signals, a collection of
signals described by a $\mdim$-dimensional manifold $\manifold
\subset \real^\dim$ can have a stable embedding in an
$\pdim$-dimensional measurement space. Moreover, the requisite
number of random measurements $\pdim$ is once again linearly
proportional to the information level (or number of degrees of
freedom) $\mdim$.

As was the case with the RIP for sparse signal processing, this
result has a number of possible implications for manifold-based
signal processing. Individual signals obeying a manifold model can
be acquired and stored efficiently using compressive measurements,
and it is unnecessary to employ the manifold model itself as part of
the compression process. Rather, the model need be used only for
signal understanding from the compressive measurements. Problems
such as Q1 (signal recovery) and Q3 (parameter estimation) can be
addressed. We have reported promising experimental results with
various classes of parametric
signals~\cite{mbwPhdThesis,davenport2007sfc}. We have also extended
Theorem~\ref{theo:manifoldjl} to the case of multiple manifolds that
are simultaneously embedded~\cite{davenport2007sfc}; this allows
both the classification of an observed object to one of several
possible models (different manifolds) and the estimation of a
parameter within that class (position on a manifold). Moreover,
collections of signals obeying a manifold model (such as multiple
images of a scene photographed from different perspectives) can be
acquired using compressive measurements, and the resulting manifold
structure will be preserved among the suite of measurement vectors
in $\real^\pdim$. We have provided empirical and theoretical support
for the use of manifold learning in the reduced-dimensional
space~\cite{hegde2007rpm}; this can dramatically simplify the
computational and storage demands on a system for processing large
databases of signals.

\subsection{Signal recovery and parameter estimation}
\label{sec:problem}

In this paper, we provide theoretical justification for the
encouraging experimental results that have been observed for
problems Q1 (signal recovery) and Q3 (parameter estimation).

To be specific, let us consider a length-$\dim$ signal $x$ that,
rather than being $\sparsity$-sparse, we assume lives on or near
some known $\mdim$-dimensional manifold $\manifold \subset
\real^\dim$. From a collection of measurements
$$y = \proj x + \noise,$$
where $\proj$ is a random $\pdim \times \dim$ matrix and $\noise \in
\real^\pdim$ is an additive noise vector, we would like to recover
either $x$ or a parameter $\theta$ that generates $x$.

For the signal recovery problem, we will consider the following as a
method for estimating $x$:
\begin{equation}
\widehat{x} = \arg\min_{x' \in \manifold} \norm{y-\proj
x'},\label{eq:csmrecover}
\end{equation}
supposing here and elsewhere that the minimum is uniquely defined.
We also let $x^\ast$ be the optimal ``nearest neighbor'' to $x$ on
$\manifold$, i.e.,
\begin{equation}
x^\ast = \arg\min_{x' \in \manifold} \norm{x-x'}. \label{eq:csmopt}
\end{equation}
To consider signal recovery successful, we would like to guarantee
that $\norm{x-\widehat{x}}$ is not much larger than
$\norm{x-x^\ast}$.

For the parameter estimation problem, where we presume $x \approx
x_\theta$ for some $\theta \in \Theta$, we propose a similar method
for estimating $\theta$ from the compressive measurements:
\begin{equation}
\widehat{\theta} = \arg\min_{\theta' \in \Theta} \norm{y-\proj
x_{\theta'}}.\label{eq:parmrecover}
\end{equation}
Let $\theta^\ast$ be the ``optimal estimate'' that could be obtained
using the full data $x \in \real^\dim$, i.e.,
\begin{equation}
\theta^\ast = \arg\min_{\theta' \in \Theta} \norm{x-x_{\theta'}}.
\label{eq:parmopt}
\end{equation}
(If $x = x_\theta$ exactly for some $\theta$, then $\theta^\ast =
\theta$; otherwise this formulation allows us to consider signals
$x$ that are not precisely on the manifold $\manifold$ in
$\real^\dim$. This generalization has practical relevance; a local
image block, for example, may only approximately resemble a straight
edge, which has a simple parameterization.) To consider parameter
estimation successful, we would like to guarantee that
$d_\Theta(\widehat{\theta},\theta^\ast)$ is small.

As we will see, bounds pertaining to accurate signal recovery can
often be extended to imply accurate parameter estimation as well.
However, the relationships between distance $d_\Theta$ in parameter
space and distances $d_\manifold$ and $\|\cdot\|_2$ in the signal
space can vary depending on the parametric signal model under study.
Thus, for the parameter estimation problem, our ability to provide
generic bounds on $d_\Theta(\widehat{\theta},\theta^\ast)$ will be
restricted. In this paper we focus primarily on the signal recovery
problem and provide preliminary results for the parameter estimation
problem that pertain most strongly to the case of isometric
parameterizations.

In this paper, we do not confront in depth the question of how a
recovery program such as (\ref{eq:csmrecover}) can be efficiently
solved. Some discussion of this matter is provided
in~\cite{mbwFocm}, with application-specific examples provided
in~\cite{mbwPhdThesis,davenport2007sfc}. Unfortunately, it is
difficult to propose a single general-purpose algorithm for solving
(\ref{eq:csmrecover}) in $\real^\pdim$, as even the problem
(\ref{eq:csmopt}) in $\real^\dim$ may be difficult to solve
depending on certain nuances (such as topology) of the individual
manifold. Nonetheless, iterative algorithms such as Newton's
method~\cite{WakinSPIEiam} have proved helpful in many problems to
date. Additional complications arise when the manifold $\manifold$
is non-differentiable, as may happen when the signals $x$ represent
2-D images. However, just as a multiscale regularization can be
incorporated into Newton's method for solving (\ref{eq:csmopt})
(see~\cite{WakinSPIEiam}), an analogous regularization can be
incorporated into a compressive measurement operator $\proj$ to
facilitate Newton's method for solving (\ref{eq:csmrecover})
(see~\cite{donohoECS,mbwPhdThesis}). For manifolds that lack
differentiability, additional care must be taken when applying
results such as Theorem~\ref{theo:manifoldjl}; we defer a study of
these matters to a subsequent paper.

In the following, we will consider both deterministic and
probabilistic instance-optimal bounds for signal recovery and
parameter estimation, and we will draw comparisons to the
sparsity-based CS results of Section~\ref{sec:csresults}. Our bounds
are formulated in terms of generic properties of the manifold (as
mentioned in Section~\ref{sec:mmodels}), which will vary from signal
model to signal model. In some cases, calculating these may be
possible, whereas in other cases it may not. Nonetheless, we feel
the results in this paper highlight the relative importance of these
properties in determining the requisite number of measurements.
Finally, to simplify analysis we will focus on random
orthoprojectors for the measurement operator $\proj$, although our
results may be extended to other random distributions such as the
Gaussian~\cite{mbwFocm}.

\section{A deterministic instance-optimal bound in $\ell_2$}
\label{sec:deter}

We begin by seeking a deterministic instance-optimal bound. That is,
for a measurement matrix $\proj$ that meets (\ref{eq:mmain}) for all
$x_1, x_2 \in \manifold$, we seek an upper bound for the relative
reconstruction error
$$
\frac{\norm{x-\widehat{x}}}{\norm{x-x^\ast}}
$$
that holds uniformly for all $x \in \real^\dim$. In this section we
consider only the signal recovery problem; however, similar bounds
would apply to parameter estimation. We have the following result
for the noise-free case, which applies not only to the manifolds
described in Theorem~\ref{theo:manifoldjl} but also to more general
sets.

\begin{thm}
Let $\manifold \subset \real^\dim$ be any subset of $\real^\dim$,
and let $\proj$ denote an $\pdim \times \dim$ orthoprojector
satisfying (\ref{eq:mmain}) for all $x_1,x_2 \in \manifold$. Suppose
$x \in \real^\dim$, let $y = \proj x$, and let the recovered
estimate $\widehat{x}$ and the optimal estimate $x^\ast$ be as
defined in (\ref{eq:csmrecover}) and (\ref{eq:csmopt}). Then
\begin{equation}
\frac{\norm{x-\widehat{x}}}{\norm{x-x^\ast}} \le
\sqrt{\frac{4\dim}{\pdim (1-\epsilon)^2} - 3 +
2\sqrt{\frac{\dim}{\pdim(1-\epsilon)^2}-1}}. \label{eq:kappabound}
\end{equation}
\label{thm:kappa}
\end{thm}

\noindent {\bf Proof:} See Appendix~\ref{app:kappa}.

~

As $\frac{\pdim}{\dim} \rightarrow 0$, the bound on the right hand side
of (\ref{eq:kappabound}) grows as
$\frac{2}{1-\epsilon}\sqrt{\frac{\dim}{\pdim}}$. Unfortunately, this
is not desirable for signal recovery. Supposing, for example, that
we wish to ensure $\norm{x-\widehat{x}} \le \KappaOne
\norm{x-x^\ast}$ for all $x \in \real^\dim$, then using the bound
(\ref{eq:kappabound}) we would require that $\pdim \ge \KappaTwo
\dim$ regardless of the dimension $\mdim$ of the manifold.

The weakness of this bound is a geometric necessity; indeed, the
bound itself is quite tight in general, as the following simple
example illustrates. Suppose $\dim \ge 2$ and let $\manifold$ denote
the line segment in $\real^\dim$ joining the points $(0,0,\dots,0)$
and $(1,0,0,\dots,0)$. Let $0 \le \gamma < \pi/2$ for some $\gamma$,
let $\pdim = 1$, and let the $1 \times \dim$ measurement matrix
$$
\proj = \sqrt{\dim} \, [\cos(\gamma); ~ -\!\sin(\gamma); ~ 0; ~ 0; ~
\cdots; ~0].
$$
Any $x_1 \in \manifold$ we may write as $x_1 =
(x_1(1),0,0,\dots,0)$, and it follows that $\proj x_1 = \sqrt{\dim}
\cos(\gamma)x_1(1)$. Thus for any pair $x_1,x_2 \in \manifold$, we
have
$$
\frac{\norm{\proj x_1 -\proj x_2}}{\norm{x_1 - x_2}} =
\frac{|\sqrt{\dim}\cos(\gamma)x_1(1)-\sqrt{\dim}\cos(\gamma)x_2(1)|}{|x_1(1)-x_2(1)|}
= \sqrt{\dim}\cos(\gamma).
$$
We suppose that $\cos(\gamma) < \frac{1}{\sqrt{\dim}}$ and thus
referring to equation (\ref{eq:mmain}) we have $(1-\epsilon)
=\sqrt{\dim}\cos(\gamma)$. Now, we may consider the signal $x =
(1,\tan(\pi/2-\gamma),0,0,\dots,0)$. We then have that $x^\ast =
(1,0,0,\dots,0)$, and $\norm{x-x^\ast} = \tan(\pi/2-\gamma)$. We
also have that $\proj x =
\sqrt{\dim}(\cos(\gamma)-\sin(\gamma)\tan(\pi/2-\gamma)) = 0$. Thus
$\widehat{x} = (0,0,\dots,0)$ and $\norm{x-\widehat{x}} =
\frac{1}{\cos(\pi/2-\gamma)}$, and so
$$
\frac{\norm{x-\widehat{x}}}{\norm{x-x^\ast}} =
\frac{1}{\cos(\pi/2-\gamma)\tan(\pi/2-\gamma)} =
\frac{1}{\sin(\pi/2-\gamma)} = \frac{1}{\cos(\gamma)} =
\frac{\sqrt{\dim}}{1-\epsilon}.
$$

It is worth recalling that, as we discussed in
Section~\ref{sec:csresults}, similar difficulties arise in
sparsity-based CS when attempting to establish a deterministic
$\ell_2$ instance-optimal bound. In particular, to ensure that
$\norm{x-\widehat{x}} \le \CddOne \norm{x-x_\sparsity}$ for all $x
\in \real^\dim$, it is known~\cite{CDD} that this requires $\pdim
\ge \CddTwo \dim$ regardless of the sparsity level $\sparsity$.

In sparsity-based CS, there have been at least two types of
alternative approaches. The first are the deterministic
``mixed-norm'' results of the type given in (\ref{eq:mixed1}) and
(\ref{eq:mixed2}). These involve the use of an alternative norm such
as the $\ell_1$ norm to measure the distance from the coefficient
vector $\alpha$ to its best $\sparsity$-term approximation
$\alpha_\sparsity$. While it may be possible to pursue similar
directions for manifold-modeled signals, we feel this is undesirable
as a general approach because when sparsity is no longer part of the
modeling framework, the $\ell_1$ norm has less of a natural meaning.
Instead, we prefer to seek bounds using $\ell_2$, as that is the
most conventional norm used in signal processing to measure energy
and error.

Thus, the second type of alternative bounds in sparsity-based CS
have involved $\ell_2$ bounds in probability, as we discussed in
Section~\ref{sec:csresults}. Indeed, the performance of both
sparsity-based and manifold-based CS is often much better in
practice than a deterministic $\ell_2$ instance-optimal bound might
indicate. The reason is that, for any $\proj$, such bounds consider
the {\em worst case} signal over all possible $x \in \real^\dim$.
Fortunately, this worst case is not typical. As a result, it is
possible to derive much stronger results that consider any given
signal $x \in \real^\dim$ and establish that for most random
$\proj$, the recovery error of that signal $x$ will be small.

\section{Probabilistic instance-optimal bounds in $\ell_2$}
\label{sec:prob}

For a given measurement operator $\proj$, our bound in
Theorem~\ref{thm:kappa} applies uniformly to any signal in
$\real^\dim$. However, a much sharper bound can be obtained by
relaxing the deterministic requirement.

\subsection{Signal recovery}

Our first bound applies to the signal recovery problem, and we
include the consideration of additive noise in the measurements.

\begin{thm} Suppose $x \in \real^\dim$. Let
$\manifold$ be a compact $\mdim$-dimensional Riemannian submanifold
of $\real^\dim$ having condition number $1/\condition$, volume
$\volume$, and geodesic covering regularity $R$. Fix $0 < \eps < 1$
and $0 < \rho < 1$. Let $\proj$ be a random $\pdim \times \dim$
orthoprojector, chosen independently of $x$, with
\begin{equation}
\pdim = O\left(\frac{\mdim \log(\dim \volume \regularity
\condition^{-1} \eps^{-1}) \log(1/\rho)}{\eps^2} \right).
\label{eq:bound3meas}
\end{equation}
Let $\noise \in \real^\pdim$, let $y = \proj x + \noise$, and let
the recovered estimate $\widehat{x}$ and the optimal estimate
$x^\ast$ be as defined in (\ref{eq:csmrecover}) and
(\ref{eq:csmopt}). If $\pdim \le \dim$, then with probability at
least $1-\rho$ the following statement holds:
\begin{equation}
\norm{x-\widehat{x}} \le (1 + 0.25\eps)\norm{x-x^\ast} +
(2+0.32\eps)\norm{\noise} + \frac{\eps^2 \condition}{936
\dim}.\label{eq:bound3}
\end{equation}
\label{theo:bound3}
\end{thm}

\noindent {\bf Proof:} See Appendix~\ref{app:bound3}.

~

The proof of this theorem, like that of
Theorem~\ref{theo:manifoldjl}, involves the Johnson-Lindenstrauss
Lemma. Our proof of Theorem~\ref{theo:bound3} extends the proof of
Theorem~\ref{theo:manifoldjl} by adding the points $x$ and $x^\ast$
to the finite sampling of points drawn from $\manifold$ that are
used to establish (\ref{eq:mmain}).

Let us now compare and contrast our bound with the analogous results
for sparsity-based CS. Like Theorem~\ref{theo:mixed2}, we consider
the problem of signal recovery in the presence of additive
measurement noise. Both bounds relate the recovery error
$\norm{x-\widehat{x}}$ to the proximity of $x$ to its nearest
neighbor in the concise model class (either $x_\sparsity$ or
$x^\ast$ depending on the model), and both bounds relate the
recovery error $\norm{x-\widehat{x}}$ to the amount $\norm{\noise}$
of additive measurement noise.
However, Theorem~\ref{theo:mixed2} is a deterministic bound whereas
Theorem~\ref{theo:bound3} is probabilistic, and our bound
(\ref{eq:bound3})  measures proximity to the concise model in the
$\ell_2$ norm, whereas (\ref{eq:mixed2}) uses the $\ell_1$ norm.

Our bound can also be compared with (\ref{eq:devore}), as both are
instance-optimal bounds in probability, and both use the $\ell_2$
norm to measure proximity to the concise model. However, we note
that unlike (\ref{eq:devore}), our bound (\ref{eq:bound3}) allows
the consideration of measurement noise.

Finally, we note that there is an additional term $\frac{\eps^2
\condition}{936 \dim}$ appearing on the right hand side of
(\ref{eq:bound3}). This term becomes relevant only when both
$\norm{x-x^\ast}$ and $\norm{\noise}$ are significantly smaller than
the condition number $\condition$, since $\eps^2 < 1$ and
$\frac{1}{936 \dim} \ll 1$. Indeed, in these regimes the signal
recovery remains accurate (much smaller than $\condition$), but the
quantity $\norm{x-\widehat{x}}$ may not remain strictly proportional
to $\norm{x-x^\ast}$ and $\norm{\noise}$. The bound may also be
sharpened by artificially assuming a condition number $1/\condition'
> 1/\condition$ for the purpose of choosing a number of measurements
$\pdim$ in (\ref{eq:bound3meas}). This will decrease the last term
in (\ref{eq:bound3}) as $\frac{\eps^2 \condition'}{936 \dim}$. In
the case where $\noise = 0$, it is also possible to resort to the
bound (\ref{eq:kappabound}); this bound is inferior to
(\ref{eq:bound3}) when $\norm{x-x^\ast}$ is large but ensures that
$\norm{x-\widehat{x}} \rightarrow 0$ when $\norm{x-x^\ast}
\rightarrow 0$.

\subsection{Parameter estimation}

Above we have derived a bound for the signal recovery problem, with
an error metric that measures the discrepancy between the recovered
signal $\widehat{x}$ and the original signal $x$.

However, in some applications it may be the case that the original
signal $x \approx x_{\theta^\ast}$, where $\theta^\ast \in \Theta$
is a parameter of interest. In this case we may be interested in
using the compressive measurements $y = \proj x + \noise$ to solve
the problem (\ref{eq:parmrecover}) and recover an estimate
$\widehat{\theta}$ of the underlying parameter.

Of course, these two problems are closely related. However, we
should emphasize that guaranteeing $\norm{x-\widehat{x}} \approx
\norm{x-x^\ast}$ does not automatically guarantee that
$\gdist(x_{\widehat{\theta}},x_{\theta^\ast})$ is small (and
therefore does not ensure that
$d_\Theta(\widehat{\theta},\theta^\ast)$ is small). If the manifold
is shaped like a horseshoe, for example, then it could be the case
that $x_{\theta^\ast}$ sits at the end of one arm but
$x_{\widehat{\theta}}$ sits at the end of the opposing arm. These
two points would be much closer in a Euclidean metric than in a
geodesic one.

Consequently, in order to establish bounds relevant for parameter
estimation, our concern focuses on guaranteeing that the geodesic
distance $\gdist(x_{\widehat{\theta}},x_{\theta^\ast})$ is itself
small.

\begin{thm}
Suppose $x \in \real^\dim$. Let $\manifold$ be a compact
$\mdim$-dimensional Riemannian submanifold of $\real^\dim$ having
condition number $1/\condition$, volume $\volume$, and geodesic
covering regularity $R$. Fix $0 < \eps < 1$ and $0 < \rho < 1$. Let
$\proj$ be a random $\pdim \times \dim$ orthoprojector, chosen
independently of $x$, with
\begin{equation*}
\pdim = O\left(\frac{\mdim \log(\dim \volume \regularity
\condition^{-1} \eps^{-1}) \log(1/\rho)}{\eps^2} \right).
\end{equation*}
Let $\noise \in \real^\pdim$, let $y = \proj x + \noise$, and let
the recovered estimate $\widehat{x}$ and the optimal estimate
$x^\ast$ be as defined in (\ref{eq:csmrecover}) and
(\ref{eq:csmopt}). If $\pdim \le \dim$ and if $1.16\norm{\noise} +
\norm{x-x^\ast} \le \condition/5$, then with probability at least
$1-\rho$ the following statement holds:
\begin{equation}
\gdist(\widehat{x}, x^\ast) \le (4 + 0.5\eps)\norm{x-x^\ast}
 + (4+0.64\eps)\norm{\noise} + \frac{\eps^2 \condition}{468 \dim}.
\label{eq:bound4}
\end{equation}
\label{theo:bound4}
\end{thm}

\noindent {\bf Proof:} See Appendix~\ref{app:bound4}.

~

In several ways, this bound is similar to (\ref{eq:bound3}). Both
bounds relate the recovery error to the proximity of $x$ to its
nearest neighbor $x^\ast$ on the manifold and to the amount
$\norm{\noise}$ of additive measurement noise.
Both bounds also have an additive term on the right hand side that
is small in relation to the condition number $\tau$.

In contrast, (\ref{eq:bound4}) guarantees that the recovered
estimate $\widehat{x}$ is near to the optimal estimate $x^\ast$ in
terms of geodesic distance along the manifold. Establishing this
condition required the additional assumption that $1.16\norm{\noise}
+ \norm{x-x^\ast} \le \condition/5$. Because $\condition$ relates to
the degree to which the manifold can curve back upon itself at long
geodesic distances, this assumption prevents exactly the type of
``horseshoe'' problem that was mentioned above, where it may happen
that $\gdist(\widehat{x}, x^\ast) \gg \norm{\widehat{x}-x^\ast}$.
Suppose, for example, it were to happen that $\norm{x-x^\ast}
\approx \condition$ and $x$ was approximately equidistant from both
ends of the horseshoe; a small distortion of distances under $\proj$
could then lead to an estimate $\widehat{x}$ for which
$\norm{x-\widehat{x}} \approx \norm{x-x^\ast}$ but
$\gdist(\widehat{x},x^\ast) \gg 0$. Similarly, additive noise could
cause a similar problem of ``crossing over'' in the measurement
space. Although our bound provides no guarantee in these situations,
we stress that under these circumstances, accurate parameter
estimation would be difficult (or perhaps even unimportant) in the
original signal space $\real^\dim$.

Finally, we revisit the situation where the original signal $x
\approx x_{\theta^\ast}$ for some $\theta^\ast \in \Theta$ (with
$\theta^\ast$ satisfying (\ref{eq:parmopt})), where the measurements
$y = \proj x + \noise$, and where the recovered estimate
$\widehat{\theta}$ satisfies (\ref{eq:parmrecover}). We consider the
question of whether (\ref{eq:bound4}) can be translated into a bound
on $d_\Theta(\widehat{\theta},\theta^\ast)$. As described in
Section~\ref{sec:mmodels}, in signal models where $\manifold$ is
isometric to $\Theta$, this is automatic: we have simply that
\begin{equation*}
\gdist(x_{\widehat{\theta}},x_{\theta^\ast}) =
d_\Theta(\widehat{\theta},\theta^\ast).
\end{equation*}
Such signal models are not nonexistent. Work by Donoho and
Grimes~\cite{DonohoGrimesISOMAP}, for example, has characterized a
variety of articulated image classes for which~(\ref{eq:isometry})
holds or for which $\gdist(x_{\theta_1},x_{\theta_2}) = \GrimesOne
d_\Theta(\theta_1, \theta_2)$ for some constant $\GrimesOne > 0$. In
other models it may hold that
$$
\GrimesTwo \gdist(x_{\theta_1},x_{\theta_2}) \le d_\Theta(\theta_1,
\theta_2) \le \GrimesThree \gdist(x_{\theta_1},x_{\theta_2})
$$
for constants $\GrimesTwo, \GrimesThree > 0$. Each of these
relationships may be incorporated to the bound (\ref{eq:bound4}).

\section{Conclusions and future work}
\label{sec:concl}

In this paper, we have considered the tasks of signal recovery and
parameter estimation using compressive measurements of a
manifold-modeled signal.
Although these problems differ substantially from the mechanics of
sparsity-based signal recovery, we have seen a number of
similarities that arise due to the low-dimensional geometry of the
each of the concise models.
First, we have seen that a sufficient number of compressive
measurements can guarantee a stable embedding of either type of
signal family, and the requisite number of measurements scales
linearly with the information level of the signal.
Second, we have seen that deterministic instance-optimal bounds in
$\ell_2$ are necessarily weak for both problems.
Third, we have seen that probabilistic instance-optimal bounds in
$\ell_2$ can be derived that give the optimal scaling with respect
to the signal proximity to the concise model and with respect to the
amount of measurement noise.
Thus, our work supports the growing empirical evidence that
manifold-based models can be used with high accuracy in compressive
signal processing.

As discussed in Section~\ref{sec:problem}, there remain several
active topics of research. One matter concerns the problem of
non-differentiable manifolds that arise from certain classes of
articulated image models. Based on preliminary and empirical
work~\cite{mbwPhdThesis,msSmashed}, we believe that a combined
multiscale regularization/measurement process is appropriate for
such problems. However, a suitable theory should be developed to
support this. A second topic of active research concerns fast
algorithms for solving problems such as (\ref{eq:csmrecover}) and
(\ref{eq:parmrecover}). Most successful approaches to date have
combined initial coarse-scale discrete searches with iterative
Newton-like refinements. Due to the problem-specific nuances that
can arise in manifold models, it is unlikely that a single
general-purpose algorithm analogous to $\ell_1$-minimization will
emerge for solving these problems. Nonetheless, advances in these
directions will likely be made by considering existing techniques
for solving (\ref{eq:csmopt}) and (\ref{eq:parmopt}) in the native
space, and perhaps by considering the multiscale measurement
processes described above.

Finally, while we have not considered stochastic models for the
parameter $\theta$ or the noise $\noise$, it would be interesting to
consider these situations as well. A starting point for such
statistical analysis may be the constrained Cram\'{e}r-Rao Bound
formulations~\cite{stoica1998crb,moore2007ccr} in which an unknown
parameter is constrained to live along a low-dimensional manifold.
However, the appropriate approach may once again be
problem-dependent, as the nearest-neighbor estimators
(\ref{eq:parmrecover}), (\ref{eq:parmopt}) we describe can be biased
for nonlinear or non-isometric manifolds.

\section*{Acknowledgements}

The author gratefully acknowledges Rice University, Caltech, and the
University of Michigan, where he resided during portions of this
research. An early version of Theorem~\ref{thm:kappa} appeared in
the author's Ph.D. thesis~\cite{mbwPhdThesis}, under the supervision
of Richard Baraniuk. Thanks to Rich and to the Rice CS research team
for many stimulating discussions.

\appendix

\section{Geodesic covering regularity} \label{app:reg}

We briefly review the definition of geodesic covering regularity and
refer the reader to~\cite{mbwFocm} for a deeper discussion.

\begin{definition}
Let $\manifold$ be a compact Riemannian submanifold of $\real^\dim$.
Given $\sres > 0$, the {\em geodesic covering number} $G(\sres)$ of
$\manifold$ is defined as the smallest number such that there exists
a set $A$ of points on $\manifold$, $\# A = G(\sres)$, so that for
all $x \in \manifold$,
\begin{equation*}
\min_{a \in A} \gdist(x,a) \le \sres.
\end{equation*}
\end{definition}

\begin{definition}
Let $\manifold$ be a compact $\mdim$-dimensional Riemannian
submanifold of $\real^\dim$ having volume $\volume$. We say that
$\manifold$ has {\em geodesic covering regularity} $\regularity$ for
resolutions $\sres \le \sres_0$ if
\begin{equation}
G(\sres) \le \frac{\regularity^\mdim
\volume\mdim^{\mdim/2}}{\sres^\mdim} \label{eq:coveringnum}
\end{equation}
for all $0 < \sres \le \sres_0$.
\end{definition}

\section{Proof of Theorem~\ref{thm:kappa}} \label{app:kappa}

Fix $\alpha \in [1-\epsilon, 1+\epsilon]$. We consider any two
points in $w_a,w_b \in \manifold$ such that
$$
\frac{\dist{\proj w_a}{\proj w_b}}{\dist{w_a}{w_b}} = \alpha,
$$
and supposing that $x$ is closer to $w_a$, i.e.,
$$
\dist{x}{w_a} \le \dist{x}{w_b},
$$
but $\proj x$ is closer to $\proj w_b$, i.e.,
$$
\dist{\proj x}{\proj w_b} \le \dist{\proj x}{\proj w_a},
$$
we seek the maximum value that
$$
\frac{\dist{x}{w_b}}{\dist{x}{w_a}}
$$
may take. In other words, we wish to bound the worst possible
``mistake'' (according to our error criterion) between two candidate
points on the manifold whose distance is scaled by the factor
$\alpha$.

This can be posed in the form of an optimization problem
\begin{eqnarray*}
\max_{x \in \real^\dim,  w_a,w_b \in \manifold}
\frac{\dist{x}{w_b}}{\dist{x}{w_a}} &\mathrm{s.t.}&\dist{x}{w_a} \le
\dist{x}{w_b},\\
&&
\dist{\proj x}{\proj w_b} \le \dist{\proj x}{\proj w_a},\\
&&\frac{\dist{\proj w_a}{\proj w_b}}{\dist{w_a}{w_b}} = \alpha.
\end{eqnarray*}
For simplicity, we may expand the constraint set to include all
$w_a,w_b \in \real^\dim$; the solution to this larger problem is an
upper bound for the solution to the case where $w_a,w_b \in
\manifold$.

The constraints and objective function now are invariant to adding a
constant to all three variables or to a constant rescaling of all
three. Hence, without loss of generality, we set $w_a = {\bf 0}$ and
$\norm{x} = 1$. This leaves
\begin{eqnarray*}
\max_{x,w_b \in \real^\dim} \dist{x}{w_b}
&\mathrm{s.t.}&\norm{x} = 1,\\
&&\dist{x}{w_b} \ge 1,\\
&&\dist{\proj x}{\proj w_b} \le \norm{\proj x},\\
&&\frac{\norm{\proj w_b}}{\norm{w_b}} = \alpha.
\end{eqnarray*}
We may safely ignore the second constraint (because of its relation
to the objective function), and we may also square the objective
function (to be later undone).

We recall that $\proj = \sqrt{\dim/\pdim}\, \Xi$, where $\Xi$ is an
$\pdim \times \dim$ matrix having orthonormal rows. We let $\Xi'$ be
an $(\dim -\pdim) \times \dim$ matrix having orthonormal rows that
are orthogonal to the rows of $\Xi$, and we define $\proj' =
\sqrt{\dim/\pdim}\, \Xi'$. It follows that for any $x' \in
\real^\dim$,
$$
\norm{\proj x'}^2 + \norm{\proj' x'}^2 = (\dim/\pdim) \norm{x'}^2.
$$

This leads to
\begin{eqnarray*}
\max_{x,w_b \in \real^\dim} (M/N) (\dist{\proj x}{\proj w_b}^2 +
\dist{\proj' x}{\proj' w_b}^2)
\end{eqnarray*}
subject to
\begin{eqnarray*}
&& \norm{\proj x}^2 + \norm{\proj' x}^2 = N/M,
\\
&& \dist{\proj x}{\proj w_b}^2 \le \norm{\proj x}^2,\\
&& \frac{\norm{\proj w_b}^2}{\norm{\proj w_b}^2 + \norm{\proj'
w_b}^2} = (M/N) \alpha^2.
\end{eqnarray*}
The last constraint may be rewritten as
$$
\norm{\proj' w_b}^2 = \norm{\proj w_b}^2
\left(\frac{\dim}{\pdim}\frac{1}{\alpha^2} - 1\right).
$$

We note that the $\proj$ and $\proj'$ components of each vector may
be optimized separately (subject to the listed constraints) because
they are orthogonal components of that vector. Define $\beta$ to be
the value of $\norm{\proj' w_b}$ taken for the optimal solution
$w_b$. We note that the constraints refer to the norm of the vector
$\proj' w_b$ but not its direction. To maximize the objective
function, then, $\proj' w_b$ must be parallel to $\proj' x$ but with
the opposite sign. Equivalently, it must follow that
\begin{equation}
\proj' w_b = -\beta \cdot \frac{\proj' x}{\norm{\proj' x}}.
\label{eq:pf2}
\end{equation}

We now consider the second term in the objective function. From
(\ref{eq:pf2}), it follows that
\begin{eqnarray}
\dist{\proj'x}{\proj'w_b}^2 &=& \norm{\proj'x\left(1 +
\frac{\beta}{\norm{\proj'x}}\right)}^2 \nonumber \\ &=&
\norm{\proj'x}^2 \cdot \left( 1 + \frac{\beta}{\norm{\proj'x}}
\right)^2. \label{eq:pf1}
\end{eqnarray}
The third constraint also demands that
$$
\beta^2 = \norm{\proj w_b}^2
\left(\frac{\dim}{\pdim}\frac{1}{\alpha^2} - 1\right).
$$
Substituting into (\ref{eq:pf1}), we have
\begin{eqnarray*}
\dist{\proj'x}{\proj'w_b}^2 &=& \norm{\proj'x}^2 \cdot \left( 1 +
2\frac{\beta}{\norm{\proj'x}} + \frac{\beta^2}{\norm{\proj'x}^2}
\right) \\ &=& \norm{\proj'x}^2 + 2\norm{\proj'x}\norm{\proj
w_b}\sqrt{\frac{\dim}{\pdim}\frac{1}{\alpha^2} - 1} \\ && \quad +
\norm{\proj
w_b}^2\left(\frac{\dim}{\pdim}\frac{1}{\alpha^2}-1\right).
\end{eqnarray*}
This is an increasing function of $\norm{\proj w_b}$, and so we seek
the maximum value that $\norm{\proj w_b}$ may take subject to the
constraints. From the second constraint we see that $\dist{\proj
x}{\proj w_b}^2 \le \norm{\proj x}^2$; thus, $\norm{\proj w_b}$ is
maximized by letting $\proj w_b = 2\proj x$. With such a choice of
$\proj w_b$ we then have
\begin{equation*}
\dist{\proj x}{\proj w_b}^2 = \norm{\proj x}^2.
\end{equation*}
We note that this choice of $\proj w_b$ {\em also} maximizes the
first term of the objective function subject to the constraints.

We may now rewrite the optimization problem, in light of the above
restrictions:
\begin{eqnarray*}
\max_{\proj x,\proj'x} (\pdim/\dim)\left(\norm{\proj x}^2 +
\norm{\proj'x}^2 + 4\norm{\proj
x}\norm{\proj'x}\sqrt{\frac{\dim}{\pdim}\frac{1}{\alpha^2} - 1} +
4\norm{\proj x}^2\left(\frac{\dim}{\pdim}\frac{1}{\alpha^2}-1\right)\right) \\
\mathrm{s.t.} \norm{\proj x}^2 + \norm{\proj'x}^2 =
\frac{\dim}{\pdim}.
\end{eqnarray*}
We now seek to bound the maximum value that the objective function
may take. We note that the single constraint implies that
$$
\norm{\proj x}\norm{\proj'x} \le \frac{1}{2}
\left(\frac{\dim}{\pdim}\right)
$$
and that $\norm{\proj x} \le \sqrt{\dim/\pdim}$ (but because these
cannot be simultaneously met with equality, our bound will not be
tight). It follows that
\begin{eqnarray*}
 && (\pdim/\dim) \left(\norm{\proj x}^2 + \norm{\proj'x}^2 +
4\norm{\proj
x}\norm{\proj'x}\sqrt{\frac{\dim}{\pdim}\frac{1}{\alpha^2} - 1} +
4\norm{\proj
x}^2\left(\frac{\dim}{\pdim}\frac{1}{\alpha^2}-1\right)\right) \\
&\le& (\pdim/\dim)\left(\frac{\dim}{\pdim} +
2\frac{\dim}{\pdim}\sqrt{\frac{\dim}{\pdim}\frac{1}{\alpha^2} - 1} +
4\frac{\dim}{\pdim}\left(\frac{\dim}{\pdim}\frac{1}{\alpha^2}-1\right)\right) \\
&=& \frac{\dim}{\pdim}\frac{4}{\alpha^2} - 3 +
2\sqrt{\frac{\dim}{\pdim}\frac{1}{\alpha^2} - 1}.
\end{eqnarray*}

Returning to the original optimization problem (for which we must
now take a square root), this implies that
$$
\frac{\dist{x}{w_b}}{\dist{x}{w_a}} \le \sqrt{
\frac{\dim}{\pdim}\frac{4}{\alpha^2} - 3 +
2\sqrt{\frac{\dim}{\pdim}\frac{1}{\alpha^2} - 1}}
$$
for any observation $x$ that could be mistakenly paired with $w_b$
instead of $w_a$ (under a projection that scales the distance
$\dist{w_a}{w_b}$ by $\alpha$). Considering the range of possible
$\alpha$, the worst case may happen when $\alpha = (1-\epsilon)$.
\qed

\section{Proof of Theorem~\ref{theo:bound3}}
\label{app:bound3}

Following the proof of Theorem~\ref{theo:manifoldjl}
(see~\cite{mbwFocm}), we let $\eps_1  = \frac{1}{13}\eps$ and $\sres
= \frac{\eps^2 \condition}{3100 \dim}$. We let $A$ be a minimal set
of points on the manifold $\manifold$ such that, for every $x' \in
\manifold$,
\begin{equation}
\min_{a \in A} \gdist(x',a) \le \sres. \label{eq:gdist}
\end{equation}
We call $A$ the set of {\em anchor points}. From
(\ref{eq:coveringnum}) we have that $\#A \le \frac{\regularity^\mdim
\volume\mdim^{\mdim/2}}{\sres^\mdim}$. The proof also describes a
finite set of points $B \supset A$ and applies the
Johnson-Lindenstrauss Lemma to this set to conclude that
\begin{equation}
 (1-\eps_1) \norm{b_1 - b_2} \le \norm{\proj b_1 - \proj
b_2} \le (1+\eps_2) \norm{b_1 - b_2} \label{eq:bembed}
\end{equation}
holds for all $b_1, b_2 \in B$. The cardinality of the set $B$
dictates the requisite number of measurements $\pdim$ in
(\ref{eq:mmeasmain}).

For our purposes, we define a new set $B' := B \cup \{x\} \cup
\{x^\ast\}$. Noting that $\proj$ is independent of both $x$ and
$x^\ast$, we may apply the Johnson-Lindenstrauss Lemma to $B'$
instead and conclude that (\ref{eq:bembed}) holds for all $b_1, b_2
\in B'$. This new set has cardinality $|B'| \le |B| + 2$, and one
may check that this does not change the order of the number of
measurements required in (\ref{eq:mmeasmain}).

Let $\widehat{a}$ denote the anchor point nearest to $\widehat{x}$
in terms of $\ell_2$ distance in $\real^\dim$. It follows that
$\norm{\widehat{x}-\widehat{a}} \le \sres$. Since
$x,x^\ast,\widehat{a} \in B'$, we know that
$$
\compactionB{\norm{\proj x - \proj \widehat{a}}}{\norm{x -
\widehat{a}}}{\epsilon_1}
$$
and
$$
\compactionB{\norm{\proj x - \proj x^\ast}}{\norm{x -
x^\ast}}{\epsilon_1}.
$$
Also, since $\widehat{x},\widehat{a} \in \manifold$, we have from
the conclusion of Theorem~\ref{theo:manifoldjl} that
$$
\compactionB{\norm{\proj \widehat{x} - \proj
\widehat{a}}}{\norm{\widehat{x} - \widehat{a}}}{\epsilon}.
$$
Finally, notice that by definition
$$
\norm{x-x^\ast} \le \norm{x-\widehat{x}}
$$
and
$$
\norm{(\proj x+ \noise)-\proj \widehat{x}} \le \norm{(\proj x+
\noise)-\proj x^\ast}.
$$

Now, combining all of these bounds and using several applications of
the triangle inequality we have
\begin{eqnarray*}
\norm{x-\widehat{x}}
&\le& \norm{x-\widehat{a}} +
\norm{\widehat{x}-\widehat{a}} \\
&\le& \norm{x-\widehat{a}} + \sres \\
&\le& \frac{1}{1-\epsilon_1}\norm{\proj x - \proj
\widehat{a}}  + \sres \\
&\le& \frac{1}{1-\epsilon_1}\left(  \norm{\proj x - \proj
\widehat{x}} + \norm{\proj \widehat{x} - \proj \widehat{a}}
\right) + \sres \\
&\le& \frac{1}{1-\epsilon_1}\left(  \norm{\proj x - \proj
\widehat{x} + \noise} + \norm{\noise} + \norm{\proj \widehat{x} -
\proj \widehat{a}}
\right) + \sres \\
&\le& \frac{1}{1-\epsilon_1}\left(  \norm{\proj x - \proj x^\ast +
n} + \norm{\noise} + \norm{\proj \widehat{x} - \proj \widehat{a}}
\right) + \sres \\
&\le& \frac{1}{1-\epsilon_1}\left(  \norm{\proj x - \proj x^\ast} +
2\norm{\noise} + \norm{\proj \widehat{x} - \proj \widehat{a}}
\right) + \sres \\
&\le& \frac{1}{1-\epsilon_1}\left(  (1+\eps_1) \norm{x - x^\ast} +
2\norm{\noise} + \norm{\proj \widehat{x} - \proj \widehat{a}}
\right) + \sres \\
&\le& \frac{1}{1-\epsilon_1}\left(  (1+\eps_1) \norm{x - x^\ast} +
2\norm{\noise} + \sres (1+\eps)
\right) + \sres \\
&=& \frac{2\norm{\noise}}{1-\epsilon_1} +
\frac{1+\epsilon_1}{1-\epsilon_1} \norm{x-x^\ast} + \sres
\left(\frac{1+\epsilon}{1-\epsilon_1} + 1\right).\\
\end{eqnarray*}
One can check that
$$
\frac{1}{1-\epsilon_1} \le 1 + 0.16\eps,
$$
$$
\frac{1+\epsilon_1}{1-\epsilon_1} \le 1 + 0.25\eps,
$$
and
$$
\frac{1+\epsilon}{1-\epsilon_1} \le 1 + 1.31\eps.
$$
Therefore, \begin{equation*} \norm{x-\widehat{x}} \le
(2+0.32\eps)\norm{\noise} + (1 + 0.25\eps)\norm{x-x^\ast} +
\frac{\eps^2 \condition}{936 \dim}.
\end{equation*}
\qed

\section{Proof of Theorem~\ref{theo:bound4}}
\label{app:bound4}

Using a simple triangle inequality and (\ref{eq:bound3}), we have
\begin{equation}
\norm{\widehat{x} - x^\ast} \le \norm{x-\widehat{x}} +
\norm{x-x^\ast} \le (2+0.32\eps)\norm{\noise} + (2 +
0.25\eps)\norm{x-x^\ast} + \frac{\eps^2 \condition}{936 \dim}.
\label{eq:bound4a}
\end{equation}

Now, since both $\widehat{x}$ and $x^\ast$ belong to $\manifold$, we
can invoke Lemma 2.3 from~\cite{niyogi}, which states that if
$\norm{\widehat{x} - x^\ast} \le \condition/2$, then
\begin{equation}
\gdist(\widehat{x}, x^\ast) \le \condition -
\condition\sqrt{1-2\norm{\widehat{x} -
x^\ast}/\condition}.\label{eq:bound4b}
\end{equation}
(This lemma guarantees that two points separated by a small
Euclidean distance are also separated by a small geodesic distance,
and so the manifold does not ``curve back'' upon itself.) To apply
this lemma, it is sufficient to know that
$$
(2+0.32\eps)\norm{\noise} + (2 + 0.25\eps)\norm{x-x^\ast} +
\frac{\eps^2 \condition}{936 \dim} \le \condition/2,
$$
i.e., that
$$
\frac{2+0.32\eps}{2+0.25\eps}\norm{\noise} + \norm{x-x^\ast} \le
\condition \left(\frac{\frac{1}{2} - \frac{\eps^2}{936
\dim}}{2+0.25\eps}\right).
$$
For the sake of neatness, we may tighten this condition to
$1.16\norm{\noise} + \norm{x-x^\ast} \le \condition/5$, which
implies the sufficient condition above (since $\eps < 1$). Thus, if
$\norm{x-x^\ast}$ and $\norm{\noise}$ are sufficiently small (on the
order of the condition number $\condition$), then we may combine
(\ref{eq:bound4a}) and (\ref{eq:bound4b}), giving
\begin{eqnarray}
\gdist(\widehat{x}, x^\ast) &\le& \condition -
\condition\sqrt{1-\frac{2}{\condition}\left(
(2+0.32\eps)\norm{\noise} + (2 +
0.25\eps)\norm{x-x^\ast} + \frac{\eps^2 \condition}{936 \dim}\right)} \nonumber \\
&=& \condition - \condition\sqrt{1-\left(
\frac{(4+0.64\eps)}{\condition}\norm{\noise} + \frac{(4 +
0.5\eps)}{\condition}\norm{x-x^\ast} + \frac{\eps^2}{468
\dim}\right)}. \label{eq:bound4c}
\end{eqnarray}
Under the assumption that $1.16\norm{\noise} + \norm{x-x^\ast} \le
\condition/5$, it follows that
$$
0 < \frac{(4+0.64\eps)}{\condition}\norm{\noise} + \frac{(4 +
0.5\eps)}{\condition}\norm{x-x^\ast} + \frac{\eps^2}{468 \dim} < 1
$$
and so
\begin{eqnarray*}
\sqrt{1-\left(\frac{(4+0.64\eps)}{\condition}\norm{\noise} +
\frac{(4 + 0.5\eps)}{\condition}\norm{x-x^\ast} + \frac{\eps^2}{468
\dim}\right)} \\ >
1-\left(\frac{(4+0.64\eps)}{\condition}\norm{\noise} + \frac{(4 +
0.5\eps)}{\condition}\norm{x-x^\ast} + \frac{\eps^2}{468
\dim}\right).
\end{eqnarray*}
This allows us to simplify (\ref{eq:bound4c}) and gives our final
result: If $1.16\norm{\noise} + \norm{x-x^\ast} \le \condition/5$,
then
\begin{equation*}
\gdist(\widehat{x}, x^\ast) \le (4+0.64\eps)\norm{\noise} + (4 +
0.5\eps)\norm{x-x^\ast} + \frac{\eps^2 \condition}{468 \dim}.
\end{equation*}
\qed

\bibliographystyle{plain}
\bibliography{IEEEabrv,mbwParmEst08bib}

\end{document}